\documentclass[10pt,twocolumn,letterpaper]{article}

\usepackage{cvpr}
\usepackage{times}
\usepackage{epsfig}
\usepackage{graphicx}
\usepackage{amsmath}
\usepackage{amssymb}
\usepackage{soul}
\usepackage{caption}
\usepackage{color}
\usepackage{nopageno}
\usepackage{wrapfig}
\usepackage[breaklinks=true,colorlinks,bookmarks=false]{hyperref}

\cvprfinalcopy 

\def\cvprPaperID{****}

\begin{document}

\title{MixNMatch: Multifactor Disentanglement and Encoding\\for Conditional Image Generation}

%\vspace{-0.2in}
\author{ Yuheng Li ~~~~~~ Krishna Kumar Singh ~~~~~~ Utkarsh Ojha ~~~~~~ Yong Jae Lee \\
University of California, Davis
}

\twocolumn[{
\maketitle
}]

\begin{abstract}
%\vspace{-0.4in}
We present MixNMatch, a conditional generative model that learns to disentangle and encode background, object pose, shape, and texture from real images with minimal supervision, for mix-and-match image generation.  We build upon FineGAN, an unconditional generative model, to learn the desired disentanglement and image generator, and leverage adversarial joint image-code distribution matching to learn the latent factor encoders.  MixNMatch requires bounding boxes during training to model background, but requires no other supervision. Through extensive experiments, we demonstrate MixNMatch's ability to accurately disentangle, encode, and combine multiple factors for mix-and-match image generation, including sketch2color, cartoon2img, and img2gif applications.
Our code/models/demo can be found at \url{https://github.com/Yuheng-Li/MixNMatch}
\vspace{-0.2in}
\end{abstract}

\section{Introduction}
\begin{figure}[t!]
    \centering
    %\hspace{0.18in}
    \includegraphics[width=0.48\textwidth]{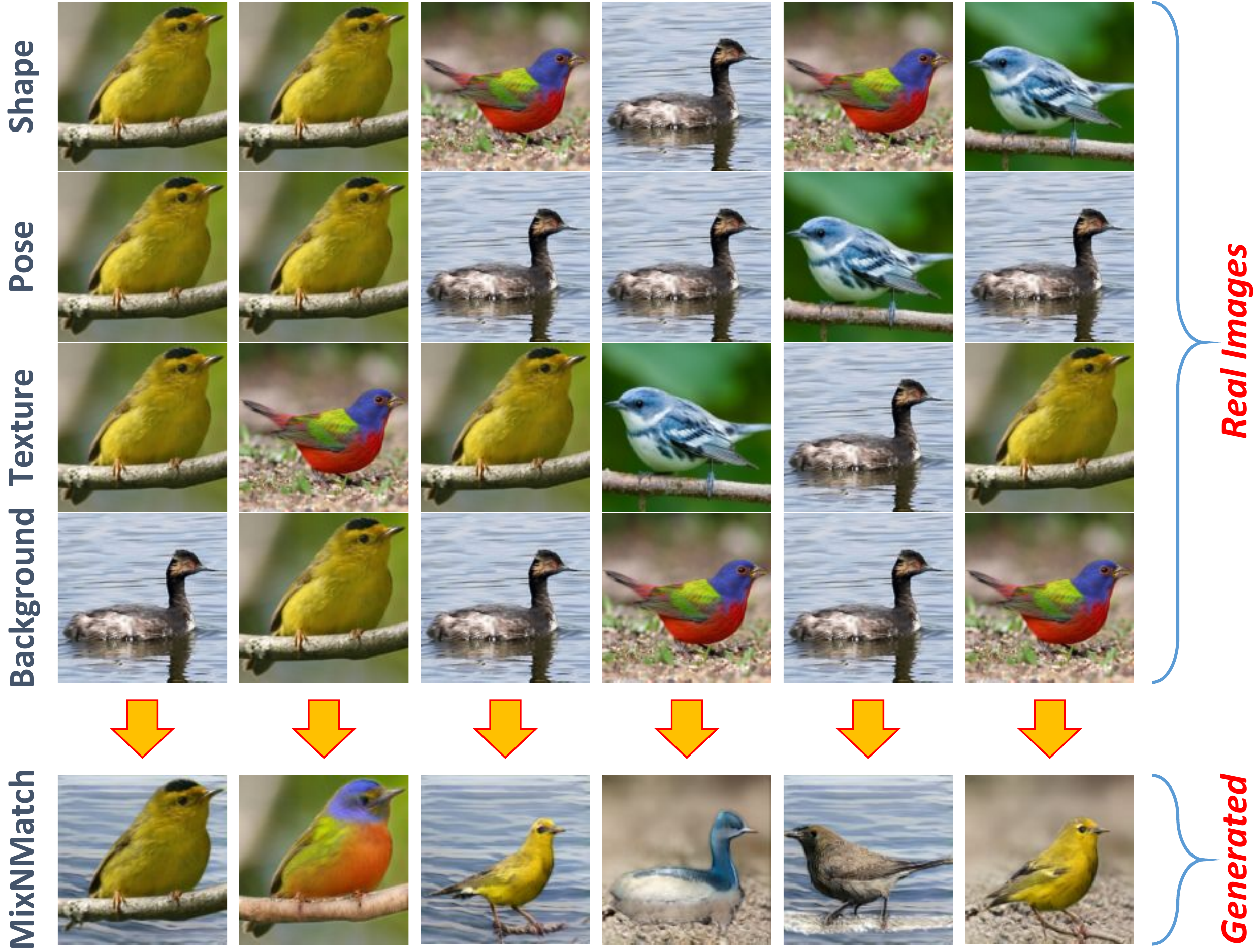}
    \caption{\textbf{Conditional mix-and-match image generation.} Our model, \emph{MixNMatch}, can disentangle and encode up to four factors---background, object pose, shape, and texture---from real reference images, and can arbitrarily combine them to generate new images. The only supervision used to train our model is bounding box annotations to model background.}
    \label{fig:concept}
    \vspace{-0.1in}
\end{figure}

%\vspace{-0.05in}
Consider the real image of the yellow bird in Figure~\ref{fig:concept} in the 1st column.  What would the bird look like in a different background, say that of the duck?  How about in a different texture, perhaps that of the rainbow textured bird in the 2nd column?  What if we wanted to keep its texture, but change its shape to that of the rainbow bird, and background and pose to that of the duck, as in the 3rd column?  How about sampling shape, pose, texture, and background from four different reference images and combining them to create an entirely new image (last column)?

\vspace{-10pt}
\paragraph{Problem.}  While research in conditional image generation has made tremendous progress~\cite{Isola-cvpr2017,zhu-iccv2017,park-cvpr2019}, no existing work can simultaneously disentangle \emph{background}, \emph{object pose}, \emph{object shape}, and \emph{object texture} with minimal supervision, so that these factors can be combined from \emph{multiple real images} for fine-grained controllable image generation. 
Learning disentangled representations with minimal supervision is an extremely challenging problem, since the underlying factors that give rise to the data are often highly correlated and intertwined.  Work that disentangle \emph{two} such factors, by taking as input two reference images e.g., one for appearance and the other for pose, do exist~\cite{huang-eccv2018,joo-cvpr18,lee-eccv18,lorenz-cvpr2019,xiao-iccv2019}. But they cannot disentangle other factors such as foreground vs.~background appearance or pose vs.~shape. Since only two factors can be controlled, these approaches cannot arbitrarily change, for example, the object's background, shape, and texture, while keeping its pose the same.  Others require strong supervision in the form of keypoint/pose/mask annotations~\cite{peng-iccv2017,balakrishnan-cvpr2018,ma-cvpr2018,esser-cvpr2018}, which limits their scalability, and still fall short of disentangling all of the four factors outlined above.

Our proposed conditional generative model, \emph{MixNMatch}, aims to fill this void.  MixNMatch learns to disentangle and encode background, object pose, shape, and texture latent factors from real images, and importantly, does so with minimal human supervision.  This allows, for example, each factor to be extracted from a different real image, and then combined together for mix-and-match image generation; see Fig.~\ref{fig:concept}.  During training, MixNMatch only requires a loose bounding box around the object to model background, but requires no other supervision for modeling the object's pose, shape, and texture.  

\vspace{-10pt}
\paragraph{Main idea.}  Our goal of mix-and-match image generation i.e., generating a single synthetic image that combines different factors from multiple real reference images, requires a framework that can simultaneously learn (1) an encoder that encodes latent factors from real images into a \emph{disentangled} latent code space, and (2) a generator that takes latent factors from the disentangled code space for image generation.  To learn the generator and the disentangled code space, we build upon FineGAN~\cite{singh-cvpr2019}, a generative model that learns to hierarchically disentangle background, object pose, shape, and texture with minimal supervision using information theory.  However, FineGAN is conditioned only on \emph{sampled latent codes}, and cannot be directly conditioned on real images for image generation.  We therefore need a way to extract latent codes that control background, object pose, shape, and texture from \emph{real images}, while preserving FineGAN's hierarchical disentanglement properties.  As we show in the experiments, a naive extension of FineGAN in which an encoder is trained to map a fake image into the codes that generated it is insufficient  to achieve disentanglement in real images due to the domain gap between real and fake images.  

To simultaneously achieve the above dual goals, we instead perform adversarial learning, whereby the joint distribution of real images and their extracted latent codes from the encoder, and the joint distribution of sampled latent codes and corresponding generated images from the generator, are learned to be indistinguishable, similar to ALI~\cite{dumoulin-iclr2017} and BiGAN~\cite{donahue-iclr2017}.  By enforcing matching joint image-code distributions, the encoder learns to produce latent codes that match the distribution of sampled codes with the desired disentanglement properties, while the generator learns to produce realistic images. To further encode a reference image's shape and pose factors with high fidelity, we augment MixNMatch with a \emph{feature mode} in which higher dimensional features of the image that preserve pixel-level structure (rather than low dimensional codes) are mapped to the learned disentangled feature space.

\vspace{-10pt}
\paragraph{Contributions.}  (1) We introduce MixNMatch, a conditional generative model that learns to disentangle and encode background, object pose, shape, and texture factors from real images with minimal human supervision. This gives MixNMatch fine-grained control in image generation, where each factor can be uniquely controlled.  MixNMatch can take as input either real reference images, sampled latent codes, or a mix of both. (2) Through various qualitative and quantitative evaluations, we demonstrate MixNMatch's ability to accurately disentangle, encode, and combine multiple factors for mix-and-match image generation.  Furthermore, we show that MixNMatch's learned disentangled representation leads to state-of-the-art fine-grained object category clustering results of real images.  (3) We demonstrate a number of interesting applications of MixNMatch including sketch2color, cartoon2img, and img2gif.

\begin{figure*}[t!]
    \centering
    %\hspace*{-8pt}
    \includegraphics[width=1\textwidth]{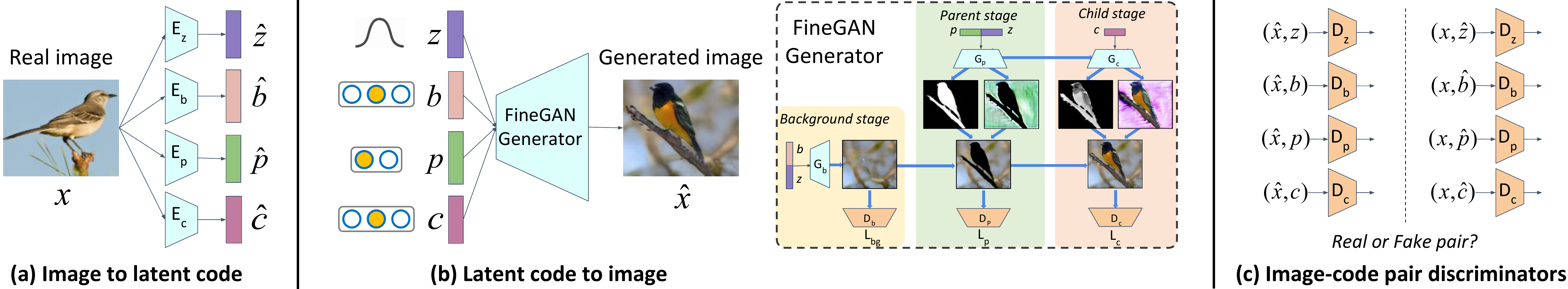}
    \caption{\textbf{MixNMatch architecture.} (a) Four different encoders, one for each factor, take a real image as input to predict the codes.  (b) Four different latent codes are sampled and fed into the FineGAN generator to hierarchically generate images. (c) Four image-code pair discriminators optimize the encoders and generator, to match their joint image-code distributions.}
    \label{fig:architecture}
    \vspace{-0.1in}
\end{figure*}

\section{Related work}

\paragraph{Conditional image generation} has various forms, including models conditioned on a class label~\cite{odena-icml2017,miyato-iclr2018,brock-iclr2019} or text input~\cite{reed-icml2016,stackgan2,xu-cvpr2018,yin-cvpr2019}.  A lot of work focuses on image-to-image translation, where an image from one domain is mapped onto another domain e.g.,~\cite{Isola-cvpr2017,zhu-iccv2017,park-cvpr2019}. However, these methods typically lack the ability to explicitly disentangle the factors of variation in the data.  Those that do learn disentangled representations focus on specific categories like faces/humans~\cite{tran-cvpr2017,peng-iccv2017,bao-cvpr2018,pumarola-eccv2018,balakrishnan-cvpr2018,ma-cvpr2018} or require clearly defined domains (e.g., pose vs.~identity or style/attribute vs.~content)~\cite{joo-cvpr18,huang-eccv2018,lee-eccv18,gonzalez-nips2018,liu-nips2018,xiao-iccv2019}.  In contrast, MixNMatch is not specific to any object category, and does not require clearly defined domains as it disentangles multiple factors of variation within a single domain (e.g., natural images of birds). Moreover, unlike most unsupervised methods which can disentangle only two factors like shape and appearance~\cite{li-ijcai2018,shu-eccv2018, lorenz-cvpr2019}, MixNMatch can disentangle four (background, object shape, pose, and texture).

\vspace{-10pt}
\paragraph{Disentangled representation learning} aims to disentangle the underlying factors that give rise to real world data~\cite{chen-nips16,yan-eccv16,xing-cvpr2018,li-ijcai2018,shu-eccv2018,tulyakov-cvpr18,hu-cvpr18,karras-cvpr2019,lorenz-cvpr2019}. Most unsupervised methods are limited to disentangling at most two factors like shape and texture~\cite{li-ijcai2018,shu-eccv2018}.  Others require strong supervision in the form of edge/keypoint/mask annotations or detectors~\cite{peng-iccv2017,balakrishnan-cvpr2018,ma-cvpr2018,esser-cvpr2018}, or rely on video to automatically acquire identity labels~\cite{denton-nips2017,joo-cvpr18,xiao-iccv2019}.  Our most related work is FineGAN~\cite{singh-cvpr2019}, which leverages information theory~\cite{chen-nips16} to disentangle background, object pose, shape, and texture with minimal supervision.  However, it is conditioned only on sampled latent codes, and thus cannot perform image translation.  We build upon this work to enable conditioning on real images.  Importantly, we show that a naive extension is insufficient to achieve disentanglement in real images. We also improve the quality of our model's image generations to preserve instance specific details from the reference images.  Since MixNMatch is directly conditioned on images, its learned representation leads to better disentanglement and fine-grained clustering of real images.

\vspace{-0.04in}
\section{Approach}
\vspace{-0.02in}

Let $\mathcal{I} = \{x_1,\dots,x_N\}$ be an unlabeled image collection of a single object category (e.g., birds). Our goal is to learn a conditional generative model, MixNMatch, which simultaneously learns to (1) encode background, object pose, shape, and texture factors associated with images in $\mathcal{I}$ into a disentangled latent code space (i.e., where each factor is uniquely controlled by a code), and (2) generate high quality images matching the true data distribution $P_{data}(x)$ by combining latent factors from the disentangled code space.

We first briefly review FineGAN~\cite{singh-cvpr2019}, from which we base our generator.  We then explain how to train our model to disentangle and encode background, object pose, shape, and texture from real images, so that it can combine different factors from different real reference images for mix-and-match image generation.  Lastly, we introduce how to augment our model to preserve object shape and pose information from a reference image with high fidelity (i.e., at the pixel-level).

\subsection{Background: FineGAN}
\label{sec:background}

FineGAN~\cite{singh-cvpr2019} takes as input four randomly sampled latent codes ($z$, $b$, $c$, $p$) to hierarchically generate an image in three stages (see Fig.~\ref{fig:architecture} (b) right): (1) a background stage where the model only generates the background, conditioned on latent one-hot background code $b$; (2) a parent stage where the model generates the object's shape and pose, conditioned on latent one-hot parent code $p$ as well as continuous code $z$, and stitches it to the existing background image; and (3) a child stage where the model fills in the object's texture, conditioned on latent one-hot child code $c$. In both the parent and child stages, FineGAN automatically generates masks (without any mask supervision) to capture the appropriate shape and texture details.

To disentangle the background, it relies on object bounding boxes (e.g., acquired through an object detector).  To disentangle the remaining factors of variation without any supervision, FineGAN uses information theory~\cite{chen-nips16}, and imposes constraints on the relationships between the latent codes.  Specifically, during training, FineGAN (1) constrains the sampled child codes into disjoint groups so that each group shares the same unique parent code, and (2) enforces the sampled background and child codes for each generated image to be the same.  The first constraint models the fact that some object instances from the same category share a common shape even if they have different textures (e.g., different duck species with different texture details share the same duck shape), and the second constraint models the fact that background is often correlated with specific object types (e.g., ducks typically have water as background). If we do not follow these constraints, then the generator could generate e.g.~a duck on a tree (background code \emph{b} not equal to texture code \emph{c}) or e.g.~a seagull with red texture (texture code \emph{c} not tied to a specific shape code \emph{p}). Then the discriminator would easily classify these images as fake, as they rarely exist in real images. As a result, the desired disentanglement will not be learned.  It is also important to note that the parent code $p$ controls viewpoint/pose invariant \emph{3D shape} of an object (e.g., duck vs.~seagull shape) as the number of unique $p$ codes is typically set to be much smaller (e.g., 20) than the amount of 2D shape variations in the data, and this in turn forces the continuous code $z$ to control viewpoint/pose.  Critically, these factors emerge as a property of the data and the model, and not through any supervision.

FineGAN is trained with three losses, one for each stage, which use either adversarial training~\cite{goodfellow-nips2014} to make the generated image look real and/or mutual information maximization~\cite{chen-nips16} between the latent code and corresponding image so that each code gains control over the respective factor (background, pose, shape, color).  We simply denote its full loss as:
\begin{equation}
\mathcal{L}_{finegan} = \mathcal{L}_{b} + \mathcal{L}_{p}+\mathcal{L}_{c},
\end{equation}
where $\mathcal{L}_{b}$, $\mathcal{L}_{p}$, and $\mathcal{L}_{c}$ denote the losses in the background, parent, and child stages.  For more details on these losses and the FineGAN architecture, please refer to~\cite{singh-cvpr2019}.

\subsection{Paired image-code distribution matching}

Although FineGAN can disentangle multiple factors to generate realistic images, it is conditioned on sampled latent codes, and cannot be conditioned on real images.  A naive post-processing extension in which encoders that learn to map fake images to the codes that generated them is insufficient to achieve disentanglement in real images due to the domain gap between real and fake images~\cite{singh-cvpr2019}, as we show in our experiments. 

Thus, to encode disentangled representations from \emph{real images} for conditional mix-and-match image generation, we need to extract the vector $z$ (controlling object pose), $b$ (controlling  background), $p$ (controlling object shape), and $c$ (controlling object texture) codes from \emph{real images}, while preserving the hierarchical disentanglement properties of FineGAN.  For this, we propose to train four encoders, each of which predict the $z, b, p, c$ codes from real images.  Since FineGAN has the ability to disentangle factors and generate images given latent codes, we naturally resort to using it as our generator, by keeping all the losses (i.e., $L_{finegan}$) to help the encoders learn the desired disentanglement.

Specifically, for each real training image $x$, we use the corresponding encoders to extract its $z, b, p, c$ codes.  However, we cannot simply input these codes to the generator to reconstruct the image, as the model would take a shortcut and degenerate into a simple autoencoder that does not preserve FineGAN's disentanglement properties (factorization into background, pose, shape, texture), as we show in our experiments. We therefore leverage ideas from ALI~\cite{dumoulin-iclr2017} and BiGAN~\cite{donahue-iclr2017,Donahue-NeurIPS2019} to help the encoders learn the \emph{inverse mapping}; i.e., a projection from real images into the code space, in a way that maintains the desired disentanglement properties. 

The key idea is to perform adversarial learning~\cite{goodfellow-nips2014,donahue-iclr2017,dumoulin-iclr2017}, so that the paired image-code distribution produced by the encoder $(x \sim P_{data},~\hat{y} \sim E(x))$ and the paired image-code distribution produced by the generator $(\hat{x} \sim G(y),~y \sim P_{code})$ are matched.  Here $E$ is the encoder, $G$ is the FineGAN generator, and $y$ is a placeholder for the latent codes $z, b, p, c$.    $P_{data}$ is the data (real image) distribution and $P_{code}$ is the latent code distribution.\footnote{Following FineGAN~\cite{singh-cvpr2019}: a continuous noise vector $z \sim \mathcal{N} (0,1)$; a categorical background code $b \sim \text{Cat}(K = N_b,p = 1/N_b)$; a categorical parent code $p \sim \text{Cat}(K = N_p,p = 1/N_p)$; and a categorical child code $c \sim \text{Cat}(K = N_c,p = 1/N_c)$. $N_b$, $N_p$, $N_c$ are the number of background, parent, and child categories and are set as hyperparameters.}   Formally, the input to the discriminator $D$ is an \emph{image-code pair}.  When training $D$, we set the paired real image $x$ and code $\hat{y}$ extracted from the encoder $E$ to be real, and the paired sampled input code $y$ and generated image $\hat{x}$ from the generator $G$ to be fake.  Conversely, when training $G$ and $E$, we try to fool $D$ so that the paired distributions $P_{(data,E(x))}$ and $P_{(G(y), code)}$ are indistinguishable, via a paired adversarial loss:
\vspace{-3pt}
\begin{align}
\mathcal{L}_{bi\_adv} =~ & \min_{G,E}\max_{D}~  \mathbb{E}_{x\sim P_{data}}\mathbb{E}_{\hat{y} \sim E(x)}[\log D(x,\hat{y})] \nonumber \\ 
+~ & \mathbb{E}_{y \sim P_{code}}\mathbb{E}_{\hat{x}\sim G(y)}[\log(1-D(\hat{x},y))].
\label{eqn:adv}
\end{align}
This loss will simultaneously enforce the (1) generated images $\hat{x} \sim G(y)$ to look real, and (2) extracted real image codes $\hat{y} \sim E(x)$ to capture the desired factors (i.e., pose, background, shape, appearance). Fig.~\ref{fig:architecture} (a-c) show our encoders, generator, and discriminators.

\subsection{Relaxing the latent code constraints}
\label{sec:constraints}

There is an important issue that we must address to ensure disentanglement in the extracted codes. FineGAN imposes strict code relationship constraints, which are key to inducing the desired disentanglement in an unsupervised way, but which can be difficult to realize in \emph{all} real images.

Specifically, recall from Sec.~\ref{sec:background} that these constraints impose a group of child codes to share the same unique parent code, and the background and child codes to always be the same. However, for any arbitrary real image, these strict relationships may not hold (e.g., a flying bird can have multiple different backgrounds in real images), and would thus be difficult to enforce in its extracted codes. In this case, the discriminator would easily be able to tell whether the image-code pair is real or fake (based on the code relationships), which will cause issues with learning. Moreover, it would also confuse the background $b$ and texture $c$ encoders since the background and child latent codes are always sampled to be the same ($b$ = $c$); i.e., the two encoders will essentially become identical (as they are always being asked to predict the same output as each other) and won't be able to distinguish between background and object texture.

\begin{figure*}[t!]
    \centering
    %\hspace*{-8pt}
    \includegraphics[width=\textwidth]{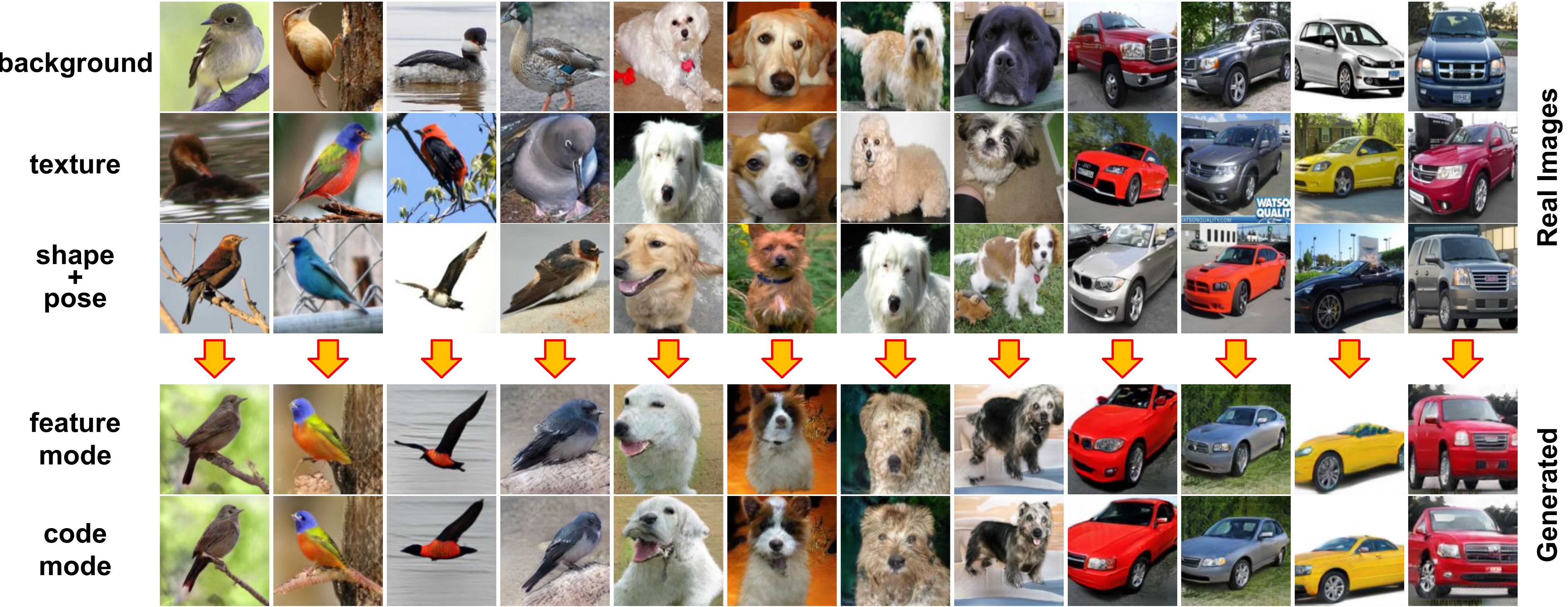}
    \caption{\textbf{Comparison between code mode \& feature mode.} Rows 1-3 are real reference images, in which we extract background $b$, texture $c$, and shape+pose $p$ \& $z$, respectively. Rows 4-5 are MixNMatch's feature mode (which accurately preserves original shape information) and code mode (which preserves shape information at a semantic level) generations.}
    \label{fig:two_modes}
    \vspace{-0.1in}
\end{figure*}

We address this issue in two ways. First, we train four separate discriminators, one for each code type.  This prevents any discriminator from seeing the other codes, and thus cannot discriminate based on the relationships between the codes.  Second, when training the encoders, we also provide as input \emph{fake} images that are generated with randomly sampled codes with the code constraints removed.  In these images, any foreground texture can be coupled with any arbitrary background ($c \neq b$) and any arbitrary shape ($c$ not tied to a particular $p$).  Specifically,  we train the encoders $E$ to predict back the sampled codes $y$ that were used to generate the corresponding fake image:
\begin{equation}
\mathcal{L}_{code\_pred} = CE(E(G(y)),y),
\label{eqn:codepred}
\end{equation}
where $CE(\cdot)$ denotes cross-entropy loss, and $y$ is a placeholder for the latent codes $ b, p, c$. (For continuous $z$, we use L1 loss.)  This loss helps to guide each encoder, and in particular the $b$ and $c$ encoders, to learn the corresponding factor.  Note that the above loss is used only to update the encoders $E$ (and not the generator $G$), as these fake images can have feature combinations that generally do not exist in the real data distribution (e.g., a duck on top of a tree).  

\subsection{Feature mode for exact shape and pose}
\label{sec:featuremode}

Thus far, MixNMatch's encoders can take in up to four different real images and encode them into $b,z,p,c$ codes which model the background, object pose, shape, and texture, respectively.  These codes can then be used by MixNMatch's generator to generate realistic images, which combine the four factors from the respective reference images.  We denote this setting as MixNMatch's \emph{code mode}.  While the generated images already capture the factors with high accuracy (see Fig.~\ref{fig:two_modes}, ``code mode''), certain image translation applications may require exact \emph{pixel-level} shape and pose alignment between a reference image and the output.

The main reason that MixNMatch in code mode cannot preserve exact pixel-level shape and pose details of a reference image is because the capacity of the latent $p$ code space, which is responsible for capturing shape, is too small to model per-instance pixel-level details (typically, tens in dimension).  The reason it must be small is because it must (roughly) match the number of unique 3D shape variations in the data (e.g., duck shape, sparrow shape, seagull shape, etc.).  In this section, we introduce MixNMatch's \emph{feature mode} to address this.  Rather than encode a reference image into a low-dimensional shape code, the key idea is to directly learn a mapping from the image to a higher-dimensional feature space that preserves the reference image's \emph{spatially-aligned} shape and pose (pixel-level) details.

\begin{wrapfigure}[8]{r}{0.3\textwidth}
	\centering
	\vspace{-15pt}
	%\hspace*{5pt}
	\includegraphics[width=0.3\textwidth]{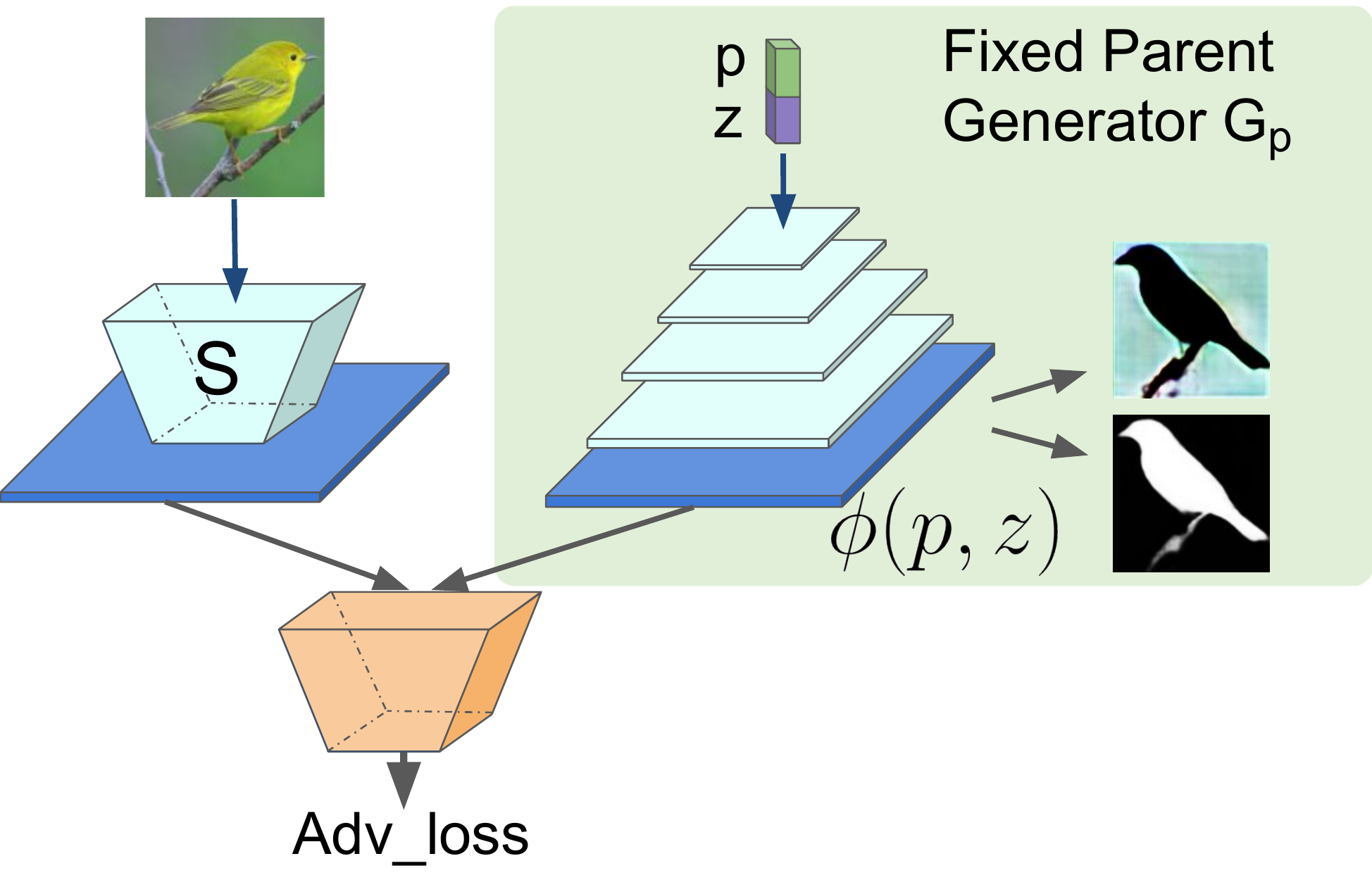}
	%\vspace{-20pt}
\end{wrapfigure}
Specifically, we take our learned parent stage generator $G_p$ (see Fig.~\ref{fig:architecture} (b)), and use it to train a new shape and pose feature extractor $S$, which takes as input a real image $x$ and outputs feature $S(x)$.  $G_p$ takes as input codes $p$ and $z$ to generate the parent stage image, which captures the object's shape. Let's denote its intermediate feature as $\phi(p, z)$. We use the standard adversarial loss~\cite{goodfellow-nips2014} to train $S$ so that the distribution of $S(x)$ matches that of $\phi(p, z)$ (i.e., only $S$ is learned and $\phi(p, z)$ is produced from the fixed pretrained $G_p$); see figure above. Ultimately, this trains $S$ to produce features that match those sampled from the $\phi(p, z)$ distribution, which already has learned to encode shape and pose. To enforce $S$ to preserve instance-specific shape and pose details of $x$ (i.e., so that the resulting generated image using $S(x)$ is spatially-aligned to $x$), we randomly sample codes $z, b, p, c$ to generate fake images using the full generator $G$, and for each fake image $G(z, b, p, c)$, we enforce an L1 loss between the feature $\phi(p, z)$ and the feature $S(G(z, b, p, c))$.

Once trained, we can use this \emph{feature mode} to extract the pixel-aligned pose and shape feature $S(x)$ from an input image $x$, and combine it with the background $b$ and texture $c$ codes extracted from (up to) two reference images, to perform conditional mix-and-match image generation.

\begin{figure*}[t!]
    \centering
    %\hspace*{-8pt}
    \includegraphics[width=\textwidth]{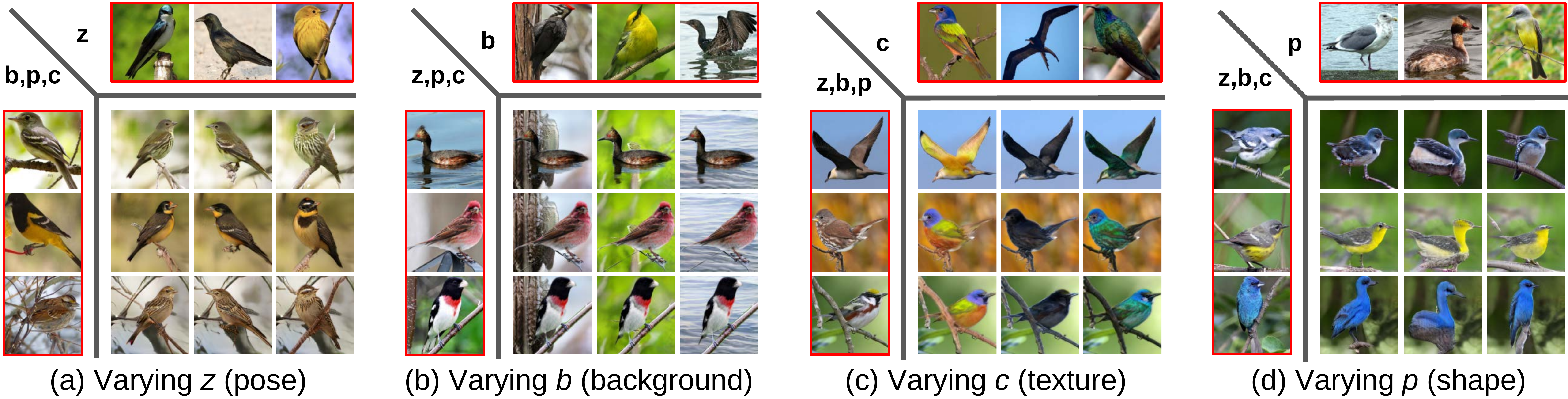}
    %\vspace{-5pt}
    \caption{\textbf{Varying a single factor.} Real images are indicated with red boxes.  For (a-d), the reference images on the left/top provide three/one factors.  The center 3x3 images are generations. For example, in (a) the top row yellow bird has an upstanding pose with its head turned to the right, and the resulting images have the same pose.}
    \label{fig:vary}
    \vspace{-0.1in}
\end{figure*}

\section{Experiments}
%\vspace{-5pt}

We evaluate MixNMatch's conditional mix-and-match image generation results, its ability to disentangle each latent factor, and its learned representation for fine-grained object clustering of real images.  We also showcase sketch2color, cartoon2img, and img2gif applications.  

\vspace{-10pt}
\paragraph{Datasets.} (1) {\bf CUB} \cite{wah-tech11}: 11,788 bird images from 200 classes; (2) {\bf Stanford Dogs}~\cite{khosla-FGVC11}: 12,000
dog images from 120 classes; (3) {\bf Stanford Cars} \cite{krause-DRR2013}: 8,144 car images from 196 classes. We set the prior latent code distributions following FineGAN~\cite{singh-cvpr2019}{\color{red}\footnotemark[1]}.  The only supervision we use is bounding boxes to model background during training.

%\vspace{-10pt}
\paragraph{Baselines.}  We compare to a number of state-of-the-art GAN, disentanglement, and clustering methods. For all methods, we use the authors' public code.  The code for SC-GAN~\cite{kazemi-wacv2018} only has the unconditional version, so we implement its BiGAN~\cite{donahue-iclr2017} variant following the paper details. 

\vspace{-10pt}
\paragraph{Implementation details.} We train and generate $128$ $\times$ $128$ images. In feature mode (2nd stage) training, $\phi(y)$ is a learned distribution from the code mode (1st stage) and may not model the entire real feature distribution (e.g., due to mode collapse).  Thus, we assume that patch-level features are better modeled, and apply a patch discriminator.  For our feature mode, since the predicted object masks are often highly accurate, we can optionally directly stitch the foreground (if only changing background) or background (if only changing texture) from the corresponding reference image. When optimizing Eqn.~\ref{eqn:adv}, we add noise to $D$ since the sampled $c$, $p$, $b$ are one hot, while predicted $\hat{c}$, $\hat{p}$, $\hat{b}$ will never be one-hot. Full training details are in the supp.

\subsection{Qualitative Results}

\paragraph{Conditional mix-and-match image generation.} We show results on CUB, Stanford Cars, and Stanford Dogs; see Fig.~\ref{fig:two_modes}. The first three rows show the background, texture, and shape + pose reference (real) images from which our model extracts $b$, $c$, and $p$ \& $z$, respectively, while the fourth and fifth rows show MixNMatch's feature mode and code mode generation results, respectively. 

Our feature mode results (4th row) demonstrate how well MixNMatch preserves shape and pose information from the reference images (3th rows), while transferring background and texture information (from 1st and 2nd rows). For example, the generated bird in the second column preserves the exact pose and shape of the bird standing on the pipe (3rd row) and transfers the brownish bark background and rainbow object texture from the 1st and 2nd row images, respectively. Our code mode results (5th row) also capture the different factors from the reference images well, though not as well as the feature mode for pose and shape. Thus, this mode is more useful for applications in which inexact instance-level pose and shape transfer is acceptable (e.g., generating a completely new instance which captures the factors at a high-level). Overall, these results highlight how well MixNMatch disentangles and encodes factors from real images, and preserves them in the generation.

Note that here we take both $z$ and $p$ from the same reference images (row 3) in order to perform a direct comparison between the code and feature modes. We next show results of disentangling all four factors, including $z$ and $p$.

\vspace{-10pt}
\paragraph{Disentanglement of factors.} Here we evaluate how well MixNMatch disentangles each factor  (background $b$, texture $c$, pose $z$, shape $p$).  Fig.~\ref{fig:vary} shows our disentanglement of each factor on CUB (results for Dogs and Cars are in the supp.). For each subfigure, the images in the top row and leftmost column (with red borders) are real reference images.  The specific factors taken from each image are indicated in the top-left corner; e.g., in (a), pose is taken from the top row, while background, shape, texture are taken from the leftmost column. Note how we can make (a) a bird change poses by varying $z$, (b) change just the background by varying $b$, (c) colorize by varying $c$, and (d) change shape by varying $p$ (e.g., see the duck example in 3rd column). As described in Sec.~\ref{sec:featuremode}, our feature mode can preserve pixel-level shape+pose information from a reference image (i.e., both \emph{p} and \emph{z} are extracted from it) in the generation. Thus, for this experiment, (b) and (c) are results of feature mode, while (a) and (d) are results of code mode. 

\begin{figure}[t!]
    \centering
    \includegraphics[width=0.4\textwidth]{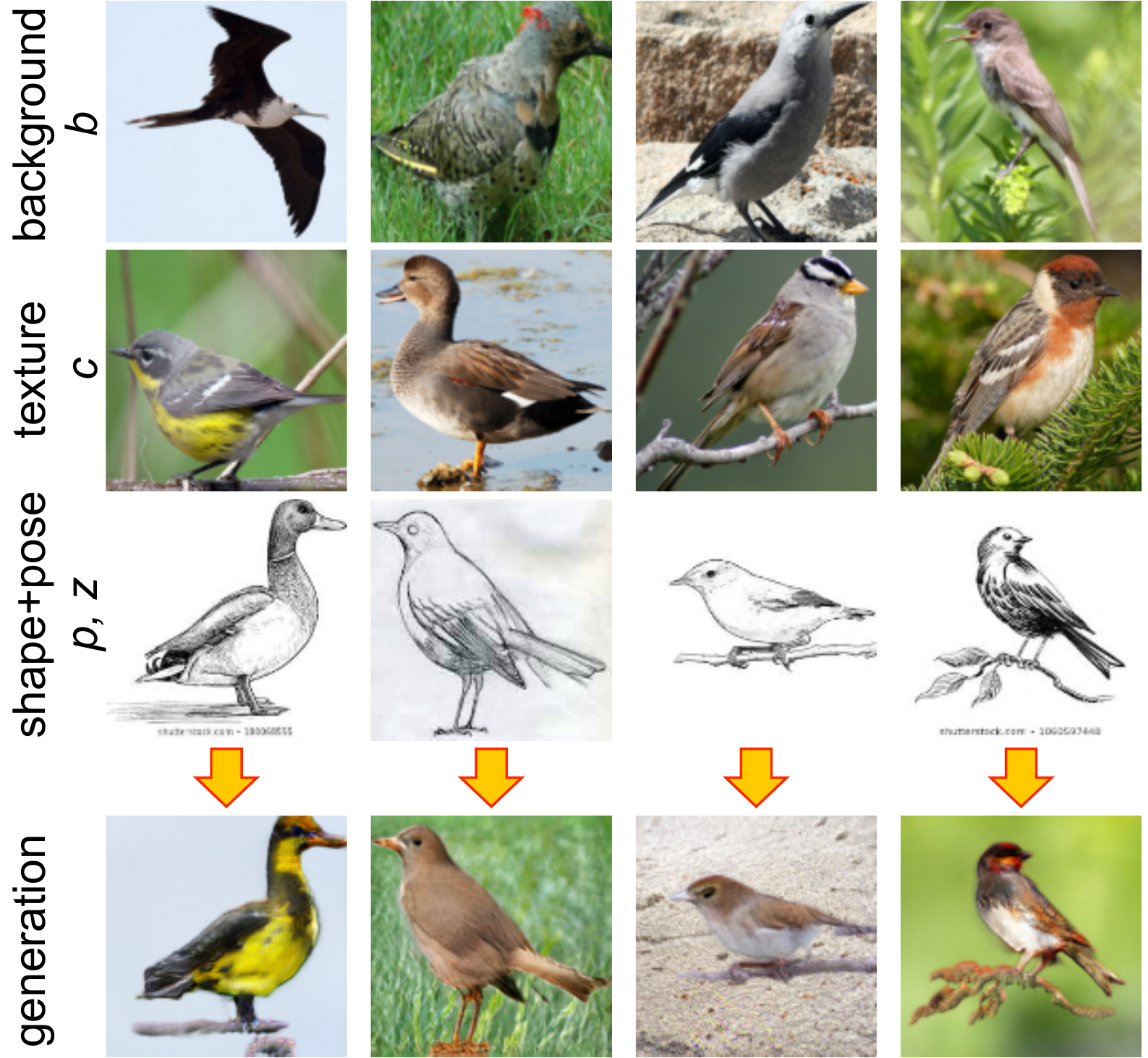}
    \caption{\textbf{sketch2color.} First three rows are real reference images. Last row shows generation results of adding background and texture to the sketch images.}
    \label{fig:sketch}
    \vspace{-0.05in}
\end{figure}

\begin{table}[t!]
    \begin{center}
        \hspace*{-5pt}
		\tabcolsep=0.1cm
		\scriptsize
		\resizebox{0.49\textwidth}{!}{
		\begin{tabular}{ l | c  c  c | c  c  c }
		    & \multicolumn{3}{c}{Inception Score} & \multicolumn{3}{|c}{FID}\\
			\hline
			 & Birds & Dogs & Cars & Birds & Dogs & Cars\\
			\hline
			Simple-GAN & 31.85 $\pm$ 0.17 & 6.75 $\pm$ 0.07 & 20.92 $\pm$ 0.14 & 16.69   & 261.85 & 33.35 \\
			InfoGAN~\cite{chen-nips16} & 47.32 $\pm$ 0.77   & 43.16 $\pm$ 0.42 & 28.62 $\pm$ 0.44 & 13.20  & 29.34 & 17.63\\
			LR-GAN~\cite{yang-iclr17} & 13.50 $\pm$ 0.20   & 10.22 $\pm$ 0.21 & 5.25 $\pm$ 0.05 & 34.91  & 54.91 & 88.80\\
			StackGANv2~\cite{stackgan2} & 43.47 $\pm$ 0.74   & 37.29 $\pm$ 0.56 & \textbf{33.69 $\pm$ 0.44} & 13.60   & 31.39 & 16.28\\
			FineGAN~\cite{singh-cvpr2019}  & \textbf{52.53} $\pm$ 0.45   & 46.92 $\pm$ 0.61 & 32.62 $\pm$ 0.37 & 11.25  & 25.66 & 16.03\\
			MixNMatch (Ours)  & 50.05 $\pm$ 0.75   & \textbf{46.97 $\pm$ 0.51} & 31.12 $\pm$ 0.62 & \textbf{9.17}  & \textbf{24.24} & \textbf{6.48} \\
			\hline
		\end{tabular}}
		\caption{\textbf{Image quality \& diversity.}  IS ($\uparrow$ better) and FID ($\downarrow$ better). MixNMatch generates diverse, high-quality images that compare favorably to state-of-the-art baselines.}
		\label{table:inception_score}
		\vspace{-0.2in}
	\end{center}
\end{table}

\begin{figure}[t!]
    \centering
    \includegraphics[width=0.4\textwidth]{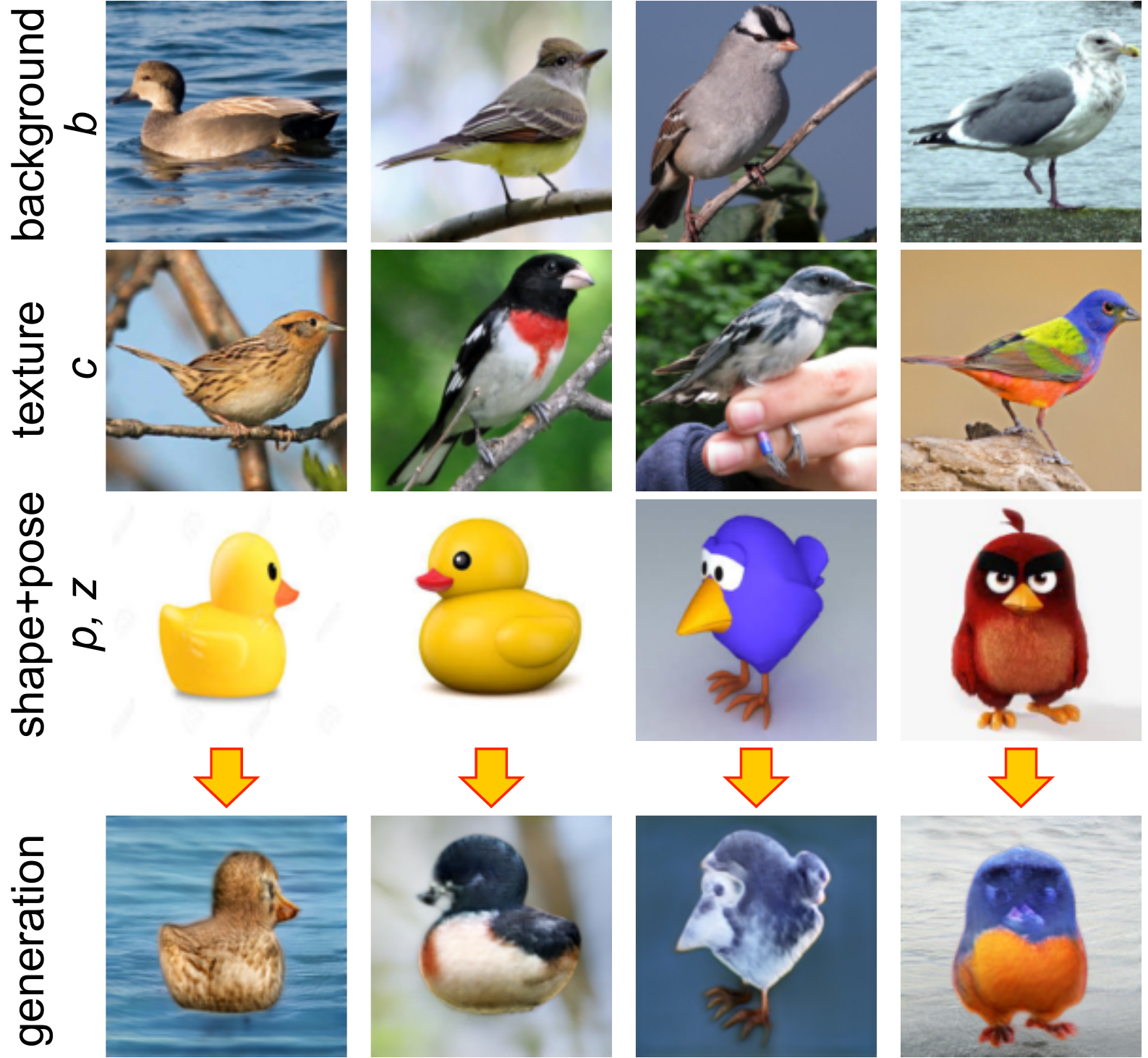}
    \caption{\textbf{cartoon2img.} MixNMatch automatically learns part semantics, \emph{without supervision}; e.g., in the 2nd column, the colors of the texture reference are accurately transferred.}
    \label{fig:cartoon}
    \vspace{-0.05in}
\end{figure}

\vspace{-10pt}
\paragraph{sketch2color / cartoon2img.} We next try adapting MixNMatch to other domains not seen during training; sketch (Fig.~\ref{fig:sketch}) and cartoon (Fig.~\ref{fig:cartoon}).  Here we use our feature mode as it can preserve pixel-level shape+pose information. Interestingly, the results indicate that MixNMatch learns \emph{part} information without supervision. For example, in Fig.~\ref{fig:cartoon} column 2, it can correctly transfer the black, white, and red part colors to the rubber duck.

\begin{figure}[t!]
    \centering
    \includegraphics[width=0.48\textwidth]{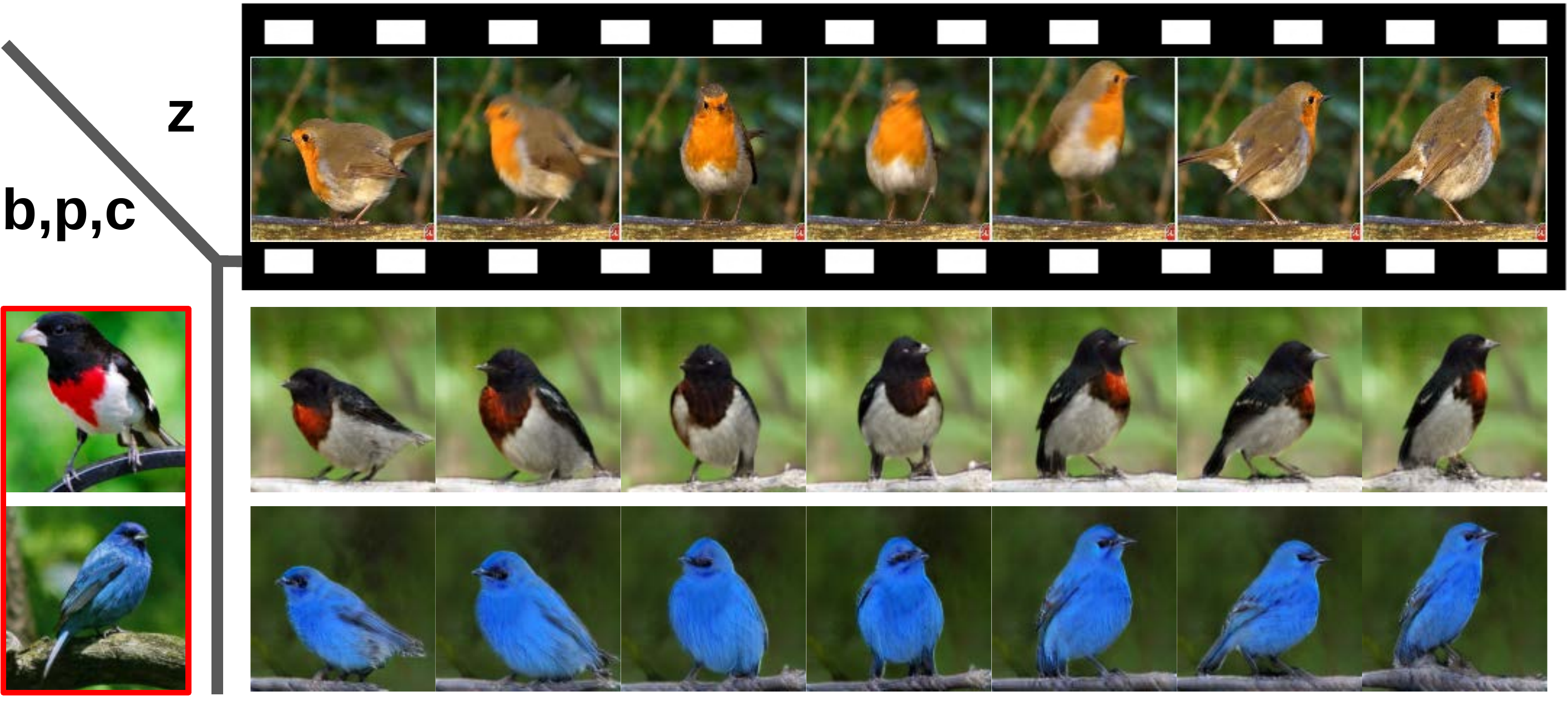}
    \caption{\textbf{image2gif.} MixNMatch can combine the pose factor $z$ from a reference video (top row), with the other factors in a static image (1st column) to animate the object.}
    \label{fig:image2gif}
    \vspace{-0.1in}
\end{figure}

%\vspace{-10pt}
\paragraph{img2gif.} MixNMatch can also be used to animate a static image; see Fig.~\ref{fig:image2gif} (code mode result) and supp. video.

\subsection{Quantitative Results}

\paragraph{Image diversity and quality.}  We compute Inception Score~\cite{salimans-nips16} and FID~\cite{FID} over 30K randomly generated images.  We condition the generation only on sampled latent codes (by sampling $z$, $p$, $c$, $b$ from their prior distributions; see Footnote 1), and not on real image inputs, for a fair comparison to the baselines.  Table~\ref{table:inception_score} shows that MixNMatch generates diverse and realistic images that are competitive to state-of-the-art unconditional GAN methods.

\vspace{-10pt}
\paragraph{Fine-grained object clustering.}  We next evaluate MixNMatch's learned representation for clustering real images into fine-grained object categories. We compare to state-of-the-art deep clustering methods: {\bf FineGAN}~\cite{singh-cvpr2019}, {\bf JULE}~\cite{yang-cvpr16}, and {\bf DEPICT}~\cite{dizaji-iccv17}, and their stronger variants~\cite{singh-cvpr2019}: {\bf JULE-Res50} and {\bf DEPICT-Large}. For evaluation metrics, we use Normalized Mutual Information ({\bf NMI})~\cite{xu-sigir03} and {\bf Accuracy}~\cite{dizaji-iccv17}, which measures the best mapping between predicted and ground truth labels. All methods cluster the same bounding box cropped images.  

To cluster real images, we use MixNMatch's $p$ (shape) and $c$ (texture) encoders as fine-grained feature extractors.  For each image, we concatenate its L2-normalized penultimate features, and run $k$-means clustering with $k$ = \# of ground-truth classes.   MixNMatch's features lead to significantly more accurate clusters than the baselines; see Table~\ref{table:nmi}.  JULE and DEPICT focus more on background and rough shape information instead of fine grained details, and thus have relatively low performance.  FineGAN performs much better, but it trains the encoders post-hoc on \emph{fake images} to repredict their corresponding latent codes (as it cannot directly condition its generator on real images)~\cite{singh-cvpr2019}.  Thus, there is a domain gap to the real image domain.  In contrast, MixNMatch's encoders are trained to extract features from both real and fake images, so it does not suffer from domain differences.

\begin{table}[t!]
	\begin{center}
	    \tabcolsep=0.15cm
		\scriptsize
		\begin{tabular}{ l | c  c  c | c  c  c }
		& \multicolumn{3}{c}{NMI} & \multicolumn{3}{|c}{Accuracy} \\
			\hline    	
			 & Birds  & Dogs & Cars & Birds  & Dogs & Cars \\	
			\hline
			JULE~\cite{yang-cvpr16}   & 0.204 & 0.142 & 0.232 & 0.045   & 0.043 & 0.046 \\
			JULE-ResNet-50~\cite{yang-cvpr16}  & 0.203 & 0.148 & 0.237 & 0.044  & 0.044 & 0.050\\
			DEPICT~\cite{dizaji-iccv17}   & 0.290 & 0.182 & 0.329 & 0.061  & 0.052 & 0.063\\
			DEPICT-Large~\cite{dizaji-iccv17}   & 0.297 & 0.183 & 0.330 & 0.061  & 0.054 & 0.062\\
			FineGAN~\cite{dizaji-iccv17}   & 0.403 & 0.233 & 0.354 & 0.126  & 0.079 & 0.078\\
			MixNMatch (Ours) & \textbf{0.422}  & \textbf{0.324} & \textbf{0.357} & \textbf{0.136}  & \textbf{0.089} & \textbf{0.079}\\
			\hline
		\end{tabular}
		\caption{\textbf{Fine-grained object clustering.}  Our approach outperforms state-of-the-art clustering methods.}
		\label{table:nmi}
		\vspace{-0.2in}
	\end{center}
\end{table}

\begin{figure}[t!]
    \centering
    \includegraphics[width=0.48\textwidth]{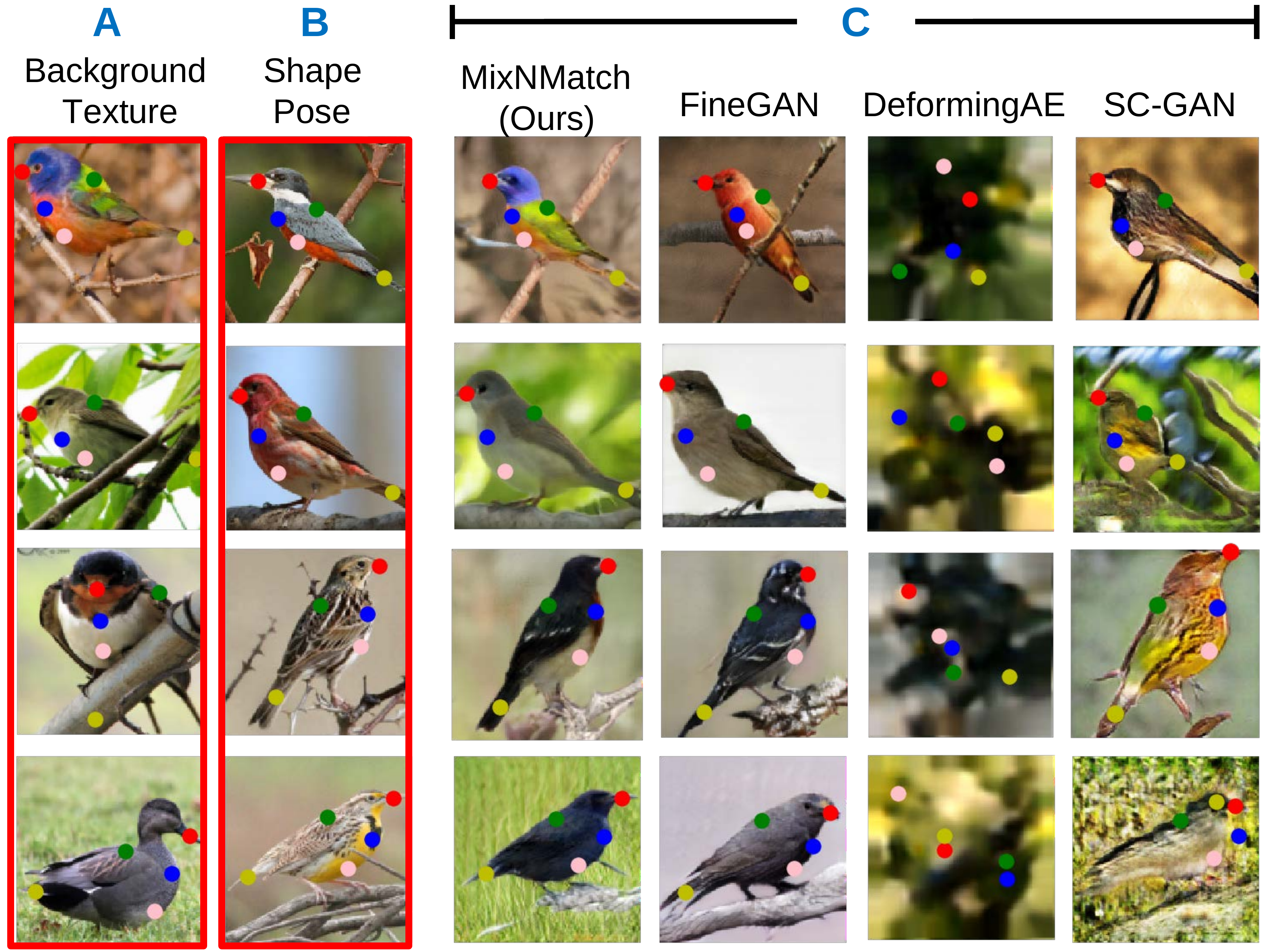}
    \caption{\textbf{Shape \& texture disentanglement.} Our approach preserves shape, texture better than strong baselines.}
    \label{fig:baseline}
    \vspace{-0.1in}
\end{figure}

\vspace{-10pt}
\paragraph{Shape and texture disentanglement.}  In order to quantitatively evaluate MixNMatch's disentanglement of shape and texture, we propose the following evaluation metric: We randomly sample 5000 image pairs (A, B) and generate new images C, which take texture and background (codes $c$, $b$) from image A, and shape and pose from image B (codes $p$, $z$).  If a model disentangles these factors well \emph{and} preserves them in the generated images, then the spatial position of part keypoints (e.g., beak, tail) in B should be close to that in C, while the texture around those keypoints in A should be similar to that in C;  see Fig.~\ref{fig:baseline}. 

To measure how well shape is preserved, we train a keypoint detector~\cite{he-iccv2017} on CUB, and use it to detect 15 keypoints in generated images C.  We then calculate the L2-distance (in x,y coordinate space) to the corresponding visible keypoints in B. To measure how well texture is preserved, for each keypoint in A and C, we first crop a 16x16 patch centered on it.  We then compute the $\chi^2$-distance between the L1-normalized color histograms of the corresponding patches in A and C.  See supp.~for more details.

Table~\ref{table:shape_color} (top) shows the results averaged over all 15 keypoints among all 5000 image triplets. We compare to \textbf{FineGAN}~\cite{singh-cvpr2019}, \textbf{SC-GAN}~\cite{kazemi-wacv2018}, a generative model that disentangles style (texture) and content (geometrical information), and \textbf{Deforming AE}~\cite{shu-eccv2018}, a generative autoencoder that disentangles shape and texture from real images via unsupervised deformation constraints. Fig.~\ref{fig:baseline} shows qualitative comparisons. Clearly, MixNMatch better disentangles and preserves shape and texture compared to the baselines. SC-GAN does not differentiate background and foreground and uses a condensed code space to model content and style, so it has difficulty transferring texture and shape accurately. Deforming AE fails because its assumption that an image can be factorized into a canonical template and a deformation field is difficult to realize in complicated shapes such as birds. FineGAN performs better, but it again is hindered by the domain gap. Finally, our feature mode has the best performance for shape disentanglement due to its ability of preserving instance-specific shape and pose details.

\begin{table}[t!]
	\begin{center}
		\scriptsize
		\begin{tabular}{ l | c  c }
			%\hline    	
			& Shape & Texture   \\	
			\hline
			Deforming AE~\cite{shu-eccv2018} & 69.97 & 0.792   \\
			SC-GAN~\cite{kazemi-wacv2018} & 32.37 &  0.641   \\
			FineGAN~\cite{singh-cvpr2019} & 21.04 & 0.602   \\
			MixNMatch (code mode) & 20.57 & \textbf{0.540}   \\
			MixNMatch (feature mode) & \textbf{16.29} & 0.565  \\
	
			\hline
			Code mode w/o paired adv loss  & 60.41 & 0.798   \\
			Code mode w/o code reprediction & 47.67 &  0.724   \\
			Code mode w/ code constraint & 26.95 & 0.601   \\
			Feature mode w/o L1 loss & 61.76 & 0.575   \\
			Feature mode w/o adv loss & 17.61 & 0.572   \\
			\hline
		\end{tabular}
		\caption{\textbf{Shape \& texture disentanglement.}  (Top) Comparisons to baselines. (Bottom) Ablation studies.  We report keypoint L2-distance and color histogram $\chi^2$-distance for measuring shape and texture disentanglement ($\downarrow$ better).}
		\label{table:shape_color}
	\end{center}
	\vspace{-0.25in}
\end{table}

\vspace{-5pt}
\paragraph{Ablation studies.} Finally, we study MixNMatch's various components: 1) \textbf{no paired image-code adversarial loss}, where we do not have Eqn.~\ref{eqn:adv}, instead we directly feed the predicted code from encoder to the generator, and apply an L1 loss between the generated and real images; 2) \textbf{without code reprediction loss}, where we do not apply Eqn.~\ref{eqn:codepred}; 3) \textbf{with code reprediction loss but with code constraints}, where during generating fake images, we keep FineGAN's code constraints; 4) \textbf{without feature mode L1 loss}, where we only apply an adversarial loss between $S(x)$ and $\phi(y)$; 5) \textbf{without feature mode adversarial loss}, where we only have the L1 loss in feature mode training. 

Table~\ref{table:shape_color} (bottom) shows that all losses are necessary in code mode training; otherwise, disentanglement cannot be learned properly. In feature mode training, both adversarial and L1 losses are helpful, as they adapt the model to the real image domain to extract precise shape + pose information. 

\vspace{-8pt}
\paragraph{Discussion.}  There are some limitations worth discussing. First, our generated background may miss large structures, as we use a patch-level  discriminator. Second, the feature mode training, depends on, and is sensitive to, how well the model is trained in the code mode. Finally, for reference images whose background and object texture are very similar, our model can fail to produce a good object mask, and thus generate an incomplete object.

\vspace{-8pt}
\paragraph{Acknowledgments.} {This work was supported in part by NSF CAREER IIS-1751206, IIS-1748387, IIS-1812850, AWS ML Research Award, Adobe Data Science Research Award, and Google Cloud Platform research credits.}

{\small
\bibliographystyle{ieee_fullname}
\bibliography{refs}
}

\clearpage

\section*{Appendix}

In this supplementary material, we first introduce some key points of our training details. Next, we elaborate on our model's feature mode (second stage) training. Then, in Sec.~\ref{sec:bg_gen}, we discuss the usage of bounding box annotations during training for background generation.  In Sec.~\ref{sec:tex_detail}, we provide details on texture disentanglement, and report shape and texture disentanglement results for all 15 keypoints for all methods.   Finally, we show more qualitative disentanglement results and discuss the video clips which further demonstrate the disentanglement ability of our model.

\section{Training details}

We optimize our model using Adam with learning rate $0.0002$, $\beta_1 = 0.5$, $\beta_2 = 0.999$ for 600 epochs. Following FineGAN~\cite{singh-cvpr2019}, we crop all the images to 1.5$\times$ of their available bounding boxes. 

As mentioned in the main paper, in our code mode (first stage) training, we use four paired discriminators to help encoders learn disentanglement. For each paired discriminator, there are two initial branches of convolution blocks which process the code and image, respectively. Then, their outputs are concatenated and fed into a series of convolution blocks to predict whether the input image-code pair is real or fake (during training, we set the image-code pair from encoders as real, and the image-code pair from generator as fake). In the code branch, we add Gaussian noise after each activation layer in order to avoid the discriminator from trivially recognizing that the one hot code in image-code pair from generator is a fake (since the encoded code from the encoders will never be one hot). Also, we update the paired discriminator using Wasserstein GAN~\cite{gulrajani-nips17} with gradient penalty.

\begin{table*}[t!]
	\begin{center}
		\begin{tabular}{ l | c c |c c| c c| c c| c c}
			%\hline
			\multicolumn{1}{l}{} & \multicolumn{2}{c}{Deforming AE~\cite{shu-eccv2018}}
			& \multicolumn{2}{c}{SC-GAN~\cite{kazemi-wacv2018}} 
			& \multicolumn{2}{c}{FineGAN~\cite{singh-cvpr2019}}
			& \multicolumn{2}{c}{ MixNMatch (c) }
			& \multicolumn{2}{c}{ MixNMatch (f) }\\	
			\hline
			 &shape&texture&shape&texture&shape&texture&shape&texture&shape&texture\\
			\hline
			back       & 75.08 & 0.816 & 27.60 & 0.679 & 16.69 & 0.637 & 16.52 & \textbf{0.561} & \textbf{13.92} & 0.584    \\
			beak       & 62.54 & 0.707 & 32.92 & 0.565 & 21.16 & 0.599 & 21.38 & \textbf{0.509} & \textbf{12.84} & 0.526    \\
			belly      & 61.58 & 0.873 & 30.86 & 0.778 & 19.56 & 0.683 & 19.51 & \textbf{0.633} & \textbf{16.92} & 0.656    \\
			breast     & 66.93 & 0.859 & 33.36 & 0.757 & 18.81 & 0.669 & 18.25 & \textbf{0.626} & \textbf{15.83} & 0.648    \\
			crown      & 81.75 & 0.773 & 32.52 & 0.631 & 19.31 & 0.614 & 18.90 & \textbf{0.550} & \textbf{12.61} & 0.564    \\
			forehead   & 70.64 & 0.759 & 29.29 & 0.572 & 18.67 & 0.570 & 18.84 & \textbf{0.495} & \textbf{11.48} & 0.510   \\
			left eye   & 66.13 & 0.809 & 27.84 & 0.586 & 17.87 & 0.540 & 17.47 & \textbf{0.481} & \textbf{12.22} & 0.508  \\
			left leg   & 70.32 & 0.800 & 34.53 & 0.573 & 26.03 & 0.585 & 24.78 & \textbf{0.508} & \textbf{22.42} & 0.537  \\
			left wing  & 68.53 & 0.809 & 34.98 & 0.714 & 25.40 & 0.609 & 24.48 & \textbf{0.572} & \textbf{22.76} & 0.612   \\
			nape       & 80.72 & 0.807 & 32.05 & 0.675 & 18.27 & 0.613 & 17.97 & \textbf{0.566} & \textbf{13.76} & 0.589   \\
			right eye  & 54.14 & 0.810 & 28.53 & 0.587 & 17.66 & 0.533 & 17.21 & \textbf{0.478} & \textbf{12.01} & 0.510   \\
			right leg  & 74.57 & 0.773 & 33.23 & 0.569 & 24.50 & 0.583 & 24.20 & \textbf{0.505} & \textbf{22.36} & 0.535   \\
			right wing & 68.99 & 0.859 & 32.28 & 0.698 & 23.43 & 0.592 & 22.84 & \textbf{0.561} & \textbf{20.00} & 0.598   \\
			tail       & 67.42 & 0.635 & 42.52 & 0.591 & 28.97 & 0.617 & 27.52 & \textbf{0.533} & \textbf{22.03} & 0.551     \\
			throat     & 80.34 & 0.792 & 33.05 & 0.641 & 19.29 & 0.596 & 18.70 & \textbf{0.527} & \textbf{13.24} & 0.554   \\
			\hline
			mean       & 69.98 & 0.792 & 32.37 & 0.641 & 21.04 & 0.602 & 20.57 & \textbf{0.540} & \textbf{16.29} & 0.565  \\
			\hline
		\end{tabular}
		\caption{\textbf{Shape \& texture disentanglement.} MixNMatch outperforms strong baselines in terms of both shape or texture disentanglement for all keypoints. (c) is code mode, (f) is feature mode.}
		\label{table:15keypoints}
	\end{center}
	\vspace{-0.25in}
\end{table*}

\section{Feature mode details}

In feature mode (second stage) training we only train a shape and pose feature extractor $S$. Concretely, we fix the trained code mode (first stage) MixNMatch generator, and treat it as a real feature distribution provider. We then randomly sample $p$ and $z$ codes from their prior code distribution (categorical and normal distribution, respectively) and also predict $p$ and $z$ using their trained encoders on randomly sampled real images with equal probability. We feed these codes into the fixed parent stage generator to get an intermediate feature $\phi(p,z)$ (we use the feature $F_p$ outputted from generator $G_p$ according to~\cite{singh-cvpr2019}). As this feature is the output of the parent stage generator, it only contains shape and pose information.  Thus, by applying an adversarial loss on the feature extractor $S$ to match the distribution of $\phi(p,z)$, we can extract shape and pose information from real images $x$.

As mentioned in the main paper, we use a patch discriminator for this feature mode (second stage) training; specifically, we use a patch size of $34$ x $34$.  Finally, in order to preserve instance-specific shape and pose details, we also generate fake images using our pretrained MixNMatch generator and compute their $\phi(p,z)$.  Then, for each fake image, we input it into the feature extractor $S$, and apply an L1 loss between the resulting feature and its $\phi(p,z)$. In summary, our loss to train $S$ is:
\begin{equation}
\mathcal{L_S}= \mathcal{L}_{adv} + \mathcal{L}_{L1} 
\end{equation}
where $\mathcal{L}_{adv}=\min\limits_{S}\max\limits_{D_S} ~\mathbb{E}_{\phi(p,z)} [\log(D_S(\phi(p,z)))] + \mathbb{E}_{x}[\log(1-D_S( S(x) ) )]$ and $\mathcal{L}_{L1}=|S(G(b,c,p,z)) - \phi(p,z)|$.  Here $D_S$ is the feature discriminator.

\section{Background modeling}\label{sec:bg_gen}

As mentioned in the main paper, we only use bounding box annotations during training to model the background.  Since we do not have any background training images without the object-of-interest (e.g., trees without bird), for each training image, we treat patches that are completely outside of the bounding box annotated (object) region as being the ``real'' background patches.  We then train the background generator $G_b$ to generate realistic background images, by applying a patch-level background discriminator $D_b$ using the adversarial loss, following~\cite{singh-cvpr2019}.  

Once our model is trained, we do not need any bounding box annotations for image generation.

\section{Shape \& texture disentanglement evaluation}\label{sec:tex_detail}
 
We first elaborate on how we evaluate texture disentanglement in Sec.~4.2 of the main paper. Recall that our goal is to take texture and background (codes $c$, $b$) from image A, shape and pose (codes $p$, $z$) from image B to generate new image C. In order to measure how well texture information is disentangled and preserved in generated image C, we first calculate 50 RGB cluster centers among 50,000 randomly sampled pixels from 1000 images (50 pixels per image) from the CUB dataset \cite{wah-tech11}.  We then fire our pretrained keypoint detector, and crop a 16x16 patch centered on each keypoint from images A and C. For each patch, we compute its histogram representation by assigning each pixel to one of the color centers. Finally, we calculate the $\chi^2$-distance between the L1-normalized color histograms of the patch in image A and corresponding patch in image C.  Since images A and B can have different poses and hence occluded parts, we only consider keypoints visible in both images.

Next, in Table~\ref{table:15keypoints}, we evaluate shape and texture disentanglement for all 15 keypoints. MixNMatch consistently outperforms the baselines for all keypoints. Our feature mode has the best performance for shape disentanglement due to its ability of preserving instance-specific shape and pose details. Our code mode model has the best performance for the texture disentanglement. One reason that the feature mode texture disentanglement result is slightly worse than that of code mode is because MixNMatch in feature mode can sometimes generate suboptimal masks (due to very similar background and object texture in the shape and pose reference images), leading to incomplete image generations.

\section{Additional results}
\begin{figure}[t!]
    \centering
    \includegraphics[width=0.48\textwidth]{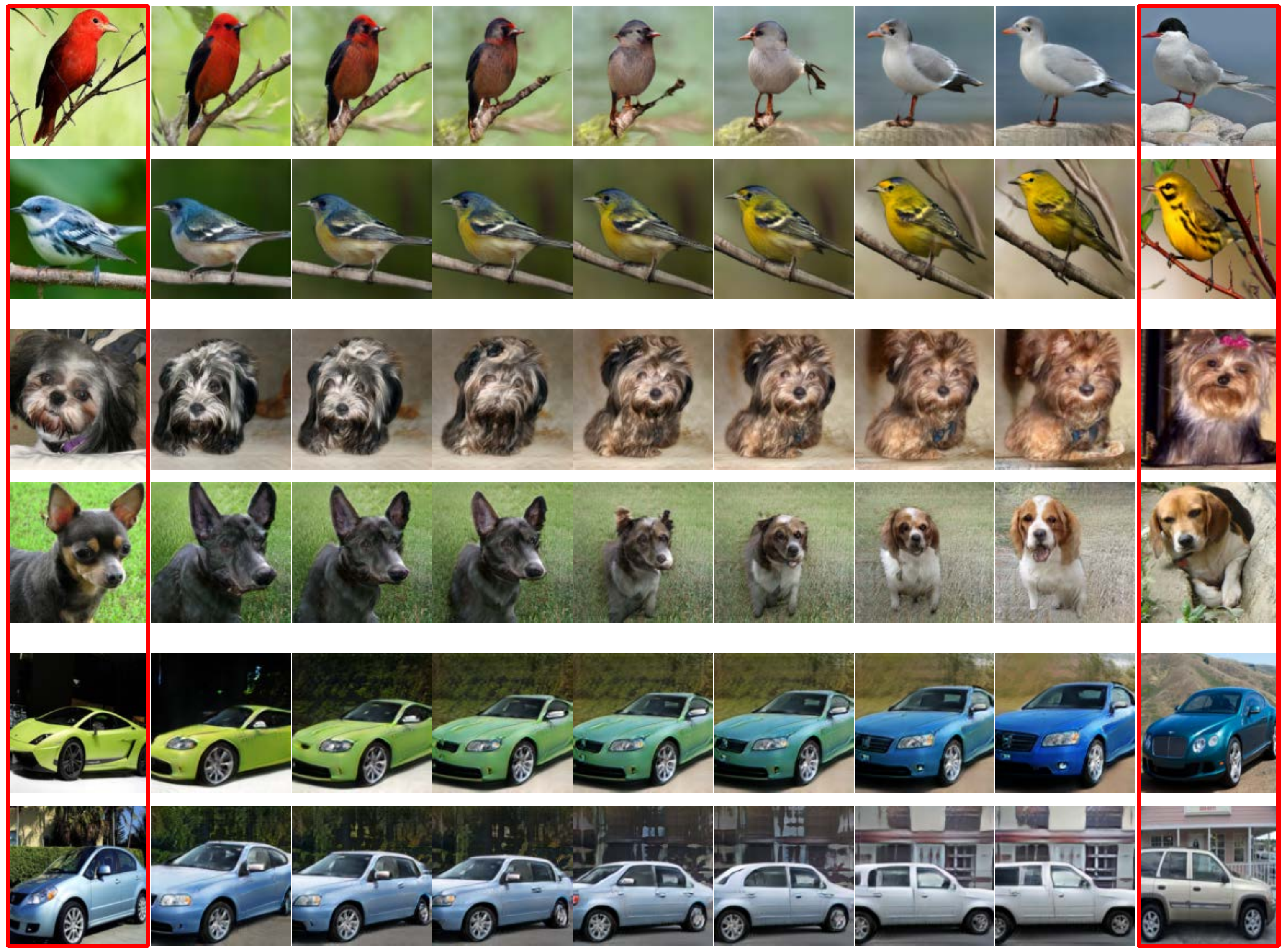}
    %\vspace{-0.1in}
    \caption{\textbf{Latent code interpolation.} Images in the red boxes are real, and intermediate images are generated by linearly interpolating codes predicted by our encoders.}
    \label{fig:interpolation}
    \vspace{-0.1in}
\end{figure}

In Fig.~\ref{fig:interpolation} we encode the $z$, $b$, $p$, $c$ codes from the two real images (first and last columns), linearly interpolate each code, and generate the interpolated images. MixNMatch produces perceptually smooth transitions for each factor, which again suggests that it has learned a highly disentangled latent space~\cite{karras-cvpr2019}.

\begin{figure*}[t!]
    \centering
    %\hspace*{-8pt}
    \includegraphics[width=\textwidth]{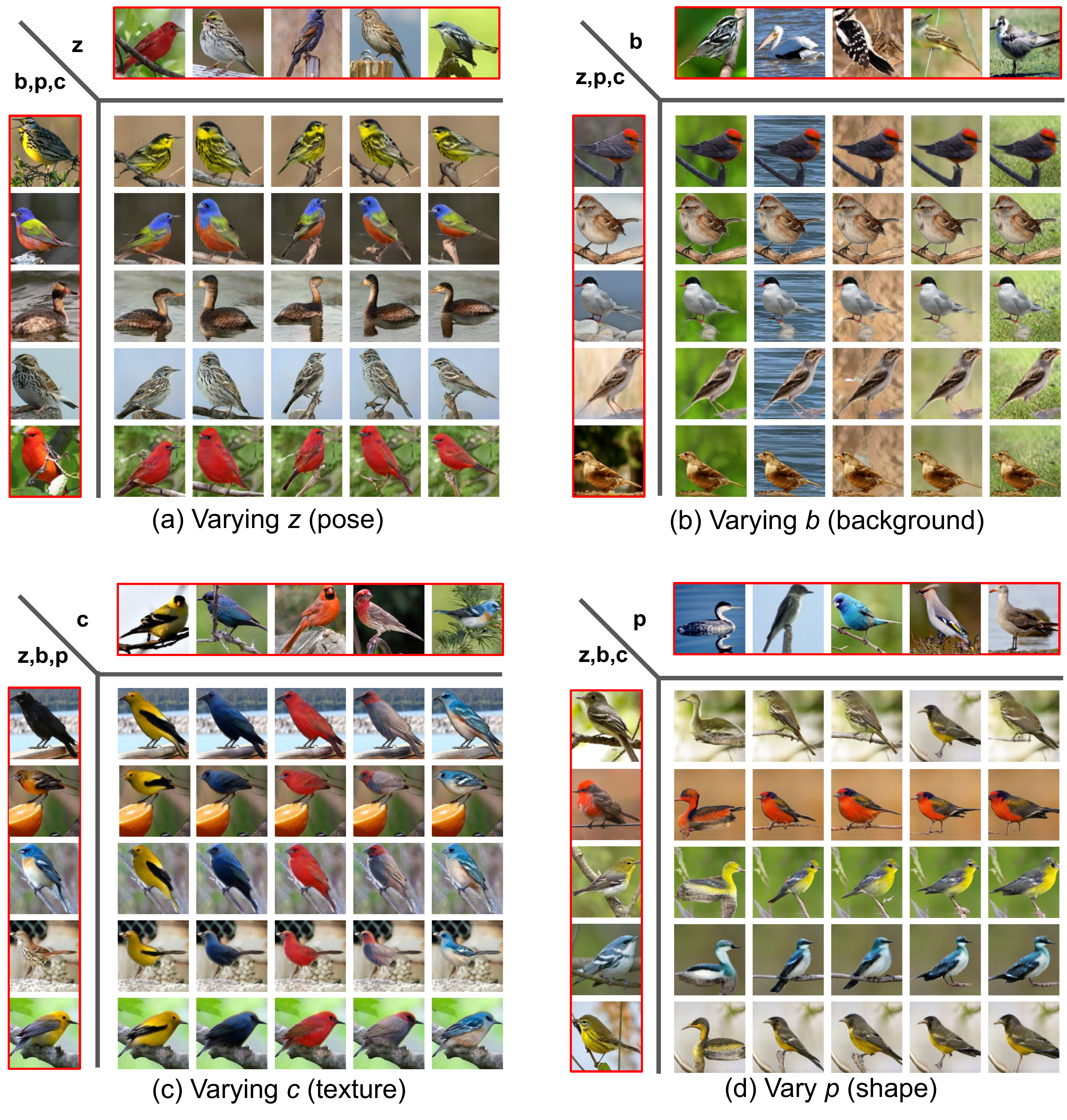}
    \caption{\textbf{Varying a single factor.} Real images are indicated with red boxes.  For (a-d), the reference images on the left/top provide three/one factors.  The center 5x5 images are generations. }
    \label{fig:bird_vary}
    \vspace{-0.1in}
\end{figure*}

\begin{figure*}[t!]
    \centering
    %\hspace*{-8pt}
    \includegraphics[width=\textwidth]{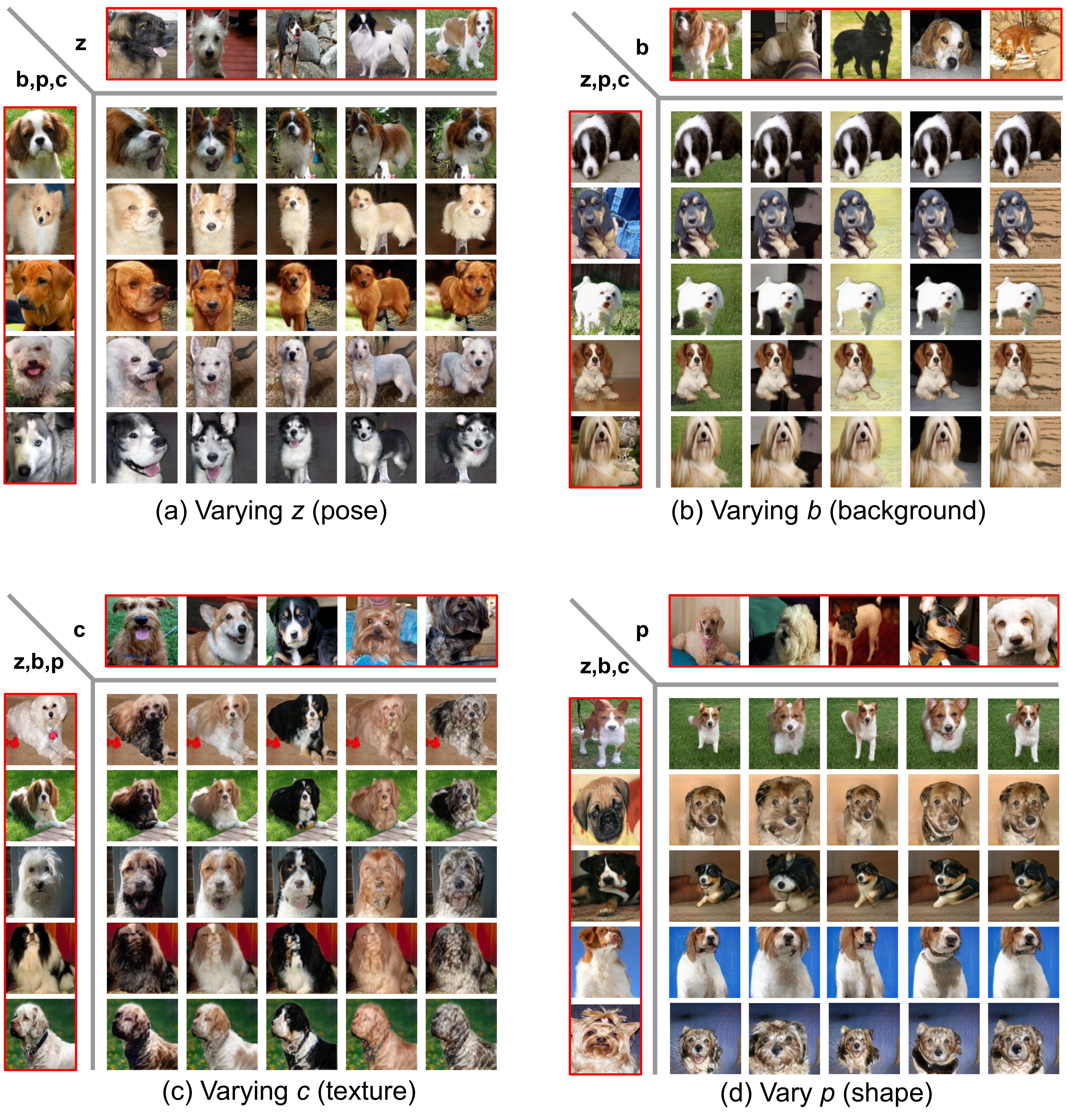}
    \caption{\textbf{Varying a single factor.} Real images are indicated with red boxes.  For (a-d), the reference images on the left/top provide three/one factors.  The center 5x5 images are generations. }
    \label{fig:dog_vary}
    \vspace{-0.1in}
\end{figure*}

\begin{figure*}[t!]
    \centering
    %\hspace*{-8pt}
    \includegraphics[width=\textwidth]{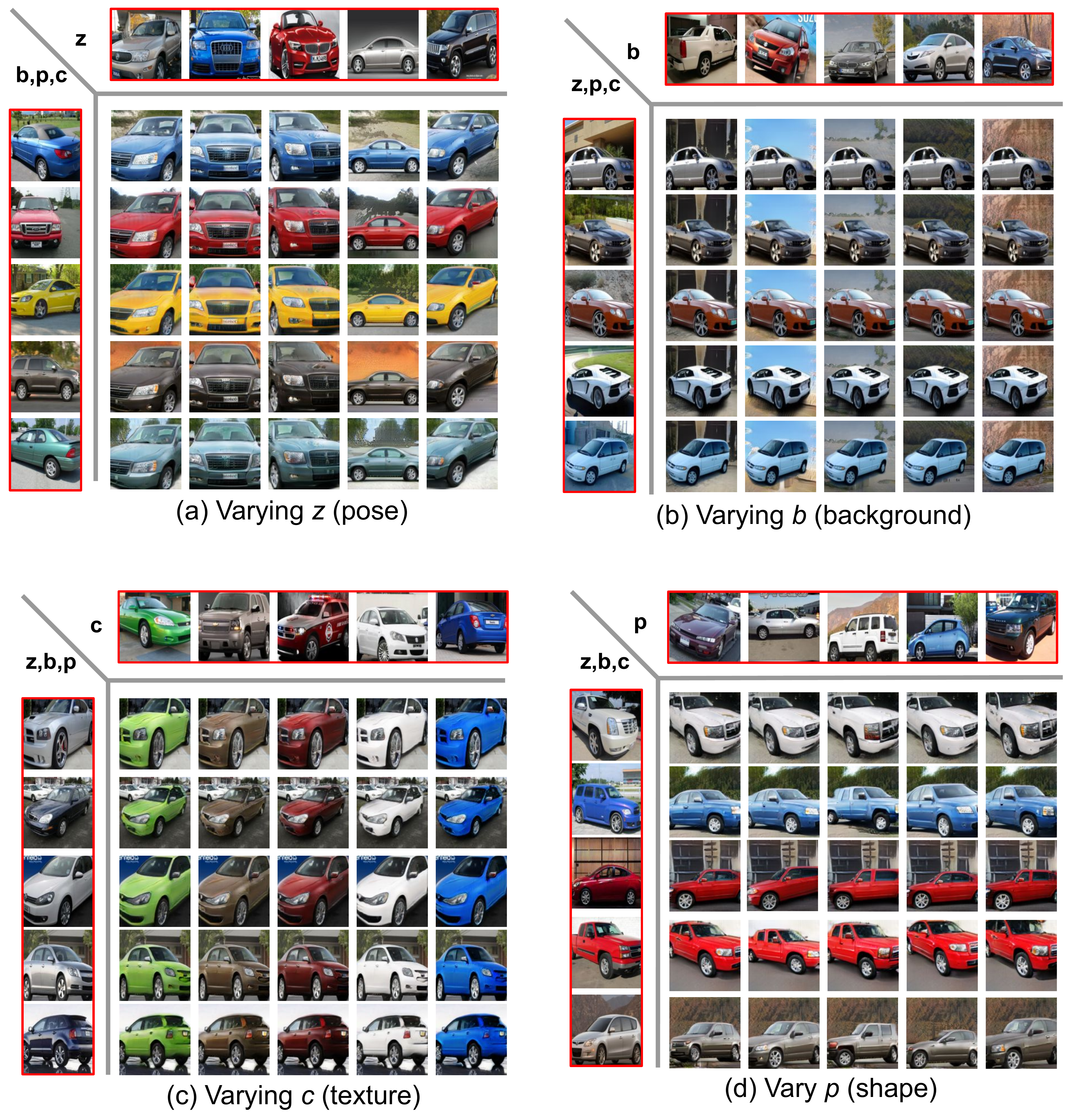}
    \caption{\textbf{Varying a single factor.} Real images are indicated with red boxes.  For (a-d), the reference images on the left/top provide three/one factors.  The center 5x5 images are generations. }
    \label{fig:car_vary}
    \vspace{-0.1in}
\end{figure*}

Figs.~\ref{fig:bird_vary}, \ref{fig:dog_vary}, and \ref{fig:car_vary} show additional disentanglement results of varying each factor for CUB, Dogs and Cars, respectively.  These results supplement Fig.~4 from the main paper.  In each sub-figure, images in the red boxes are real and we only change one factor indicated in the top left corner for generating the new images.

\section{Video results}

Finally, we include two videos demonstrating the disentanglement learned by MixNMatch. In MixNMatch.mp4, the four reference images on the top are real images which provide the four factors (background, shape, texture, and pose, respectively). The generated image is shown at the bottom. Each time we change different real reference image(s) and smoothly translate the corresponding factor.   

We also animate an object in a still image according to the  movement of a different object from a reference video.  In the two img2gif files, the frames from the reference video on the top is used to extract the $z$ vector to control object pose and location. On the left, we have a reference image from which shape, background, and texture ($p$, $b$, $c$) information are extracted. These factors are combined by MixNMatch to generate the new images at the bottom.  

Notice how our generated bird follows the pose of the reference video bird well -- e.g., it turns around and lifts its head at the end. These results clearly indicate that our model can correctly disentangle pose information from the real images. Since MixNMatch is not trained on any video data and does not use any temporal information, the generated video can be a bit sensitive and unstable in terms of the bird's shape/size.  Still, overall, each generated frame captures the factors from the respective image/video-frame very well to produce a realistic image with the corresponding properties.

\end{document}